\documentclass[conference]{IEEEtran}

\usepackage{cite}

\ifCLASSINFOpdf
  \usepackage[pdftex]{graphicx}
\else
\fi

\usepackage[cmex10]{amsmath}
\usepackage{amssymb}
\usepackage{array}

\usepackage{adjustbox}
\usepackage{footnote}

\usepackage{url}

\begin{document}
\title{A survey of robot learning from demonstrations\\for Human-Robot Collaboration}

\author{\IEEEauthorblockN{Jangwon Lee}
\IEEEauthorblockA{School of Informatics and Computing\\
Indiana University\\
Bloomington, Indiana, USA\\
Email: leejang@indiana.edu}}

\maketitle

\begin{abstract}
Robot learning from demonstration (LfD) is a research paradigm
that can play an important role in addressing the issue of scaling up robot learning.
Since this type of approach enables non-robotics experts can teach robots new knowledge
without any professional background of mechanical engineering or computer programming skills,
robots can appear in the real world even if it does not have
any prior knowledge for any tasks like a new born baby.
There is a growing body of literature that employ LfD approach for training robots.
In this paper, I present a survey of recent research in this area
while focusing on studies for human-robot collaborative tasks. 
Since there are different aspects between stand-alone tasks and collaborative tasks,
researchers should consider these differences to design collaborative robots
for more effective and natural human-robot collaboration (HRC).
In this regard, many researchers have shown an increased interest in
to make better communication framework between robots and humans
because communication is a key issue to apply LfD paradigm for human-robot collaboration. 
I thus review some recent works that focus on designing better communication channels/methods
at the first, then deal with another interesting research method, Interactive/Active learning,
after that I finally present other recent approaches tackle a more challenging problem,
learning of complex tasks, in the last of the paper.
\end{abstract}



% \fi
\IEEEpeerreviewmaketitle

\section{Introduction}
\label{sec:1}
Robot learning from demonstration (LfD) is a promising approach
that can transfer many robot prototypes remaining in research laboratories to the real world
since it typically does not require any expert knowledge of robotics technology
for teaching robots new tasks.
It thus allows end-users to teach robots what robots should do
based on their own requirements at their place.
Existing research recognizes this attractive feature of LfD,
so there is a growing body of literature that employs the theme of LfD 
for their research \cite{billard2008robot, argall2009survey, chernova2014robot}.

LfD also has been attracting a lot of interest
from researchers in the field of Human-Robot Interaction (HRI) 
because it helps robots to learn new tasks that are infeasible to be learned using pre-programming
like personal requirements as their human counterparts.
Furthermore, the HRI perspective can help to build a robot learning process more efficiently
(e.g., a human user can correct the robot's behaviors during interaction and highlight important points of the new tasks).

In order to employ the concept of LfD for human-robot collaborative tasks,
however, researchers should not only consider robot learning algorithms or techniques,
but also take into account many human-centric issues
such as the human partner's feelings and intentions during collaboration phases.
Moreover, here are a number of important differences between
the approaches that are used by HRI researchers and other robotics researchers who have different perspectives
even though both fields work under the same LfD paradigm.
One major difference is that
HRI researchers tend to be more focused on communications between humans and robots,
and thus try to extract intentions behind actions and make communications clear
while other robotics engineers are more concerned about specific techniques
like how to replicate arm trajectories from human demonstrations
for accomplishing a certain type of task.

\begin{figure}[t]
 \centering
    \includegraphics[width=0.489\textwidth]{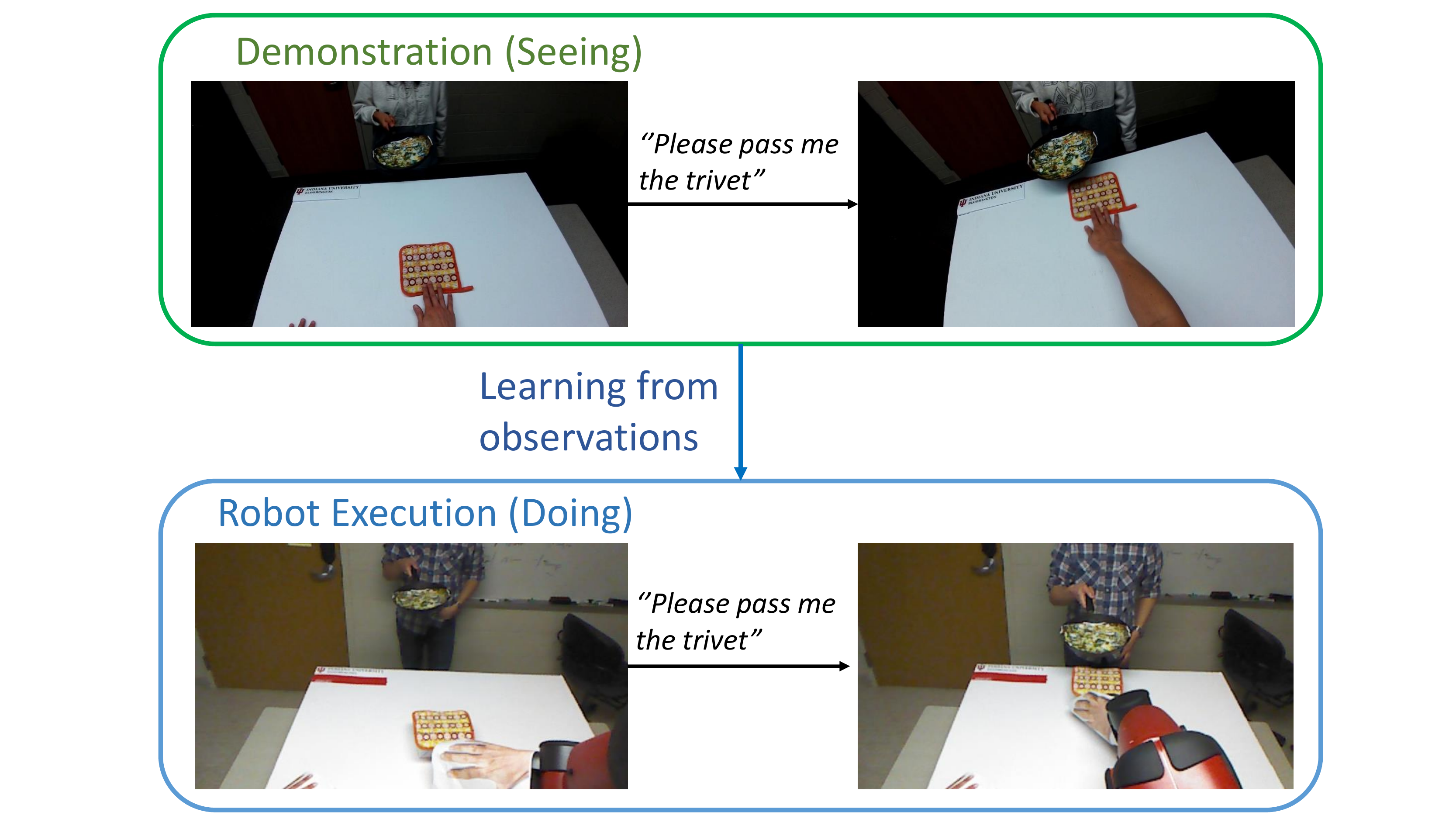}
 \caption{Robot learning from demonstration (LfD) enables robots
          automatically learn a new task from observations.
          End-users can teach robots a new task what robots should do
          without any expert knowledge of robotics technology.}
 \label{fig:overview}
\end{figure}

LfD is a broad topic ranging from various machine learning techniques
like supervised learning, reinforcement learning, and feature selection, to human factors as well.
Researchers with different backgrounds
thus employ the concept from different points of view
\cite{billard2008robot, argall2009survey, chernova2014robot}.
There are common theoretical issues in the field
such as the \emph{Correspondence Problem} that arise due to a mismatch between
the teacher's body configuration and the student's configuration \cite{nehaniv2002correspondence}
and interface issue to design user friendly interfaces for demonstration (i.e., motion-capture systems \cite{yokokohji2005motion})
in order to enable non-robotics experts to teach robots new knowledge without any difficulty.
However, I more focus on the specifics of this area
which apply this concept for human-robot collaborative tasks
that have some different issues coming from human factors
rather than to address those common issues in the field.

I begin my review by presenting an overview of LfD approaches
for human-robot collaborative tasks.
I will then go on to review research
that address communication problems between robots and humans in Section 2-A.
The approaches that employ \emph{Interactive/Active Learning} methods,
what make a teaching/learning process more interactive, 
will be addressed in the following section (Section 2-B).
In the next section, I will introduce some recent work
that attempts to teach robots a complex task rather than a single task (Section 2-C).
To conclude, I summarize this article in Section 3 with discussion.

\section{LfD for Human-Robot Collaborative tasks}
\label{sec:2}

In order to build robots that would work together side by side with humans
while sharing workplaces, many human factors should be addressed properly.
For example, researchers should handle a safety issue to prevent potential hazards by robots \cite{lasota2013developing}
and they also need to consider a human partner's mental states (i.e., feelings, desires, intents, etc.)
to make them feel more comfortable with robot co-workers \cite{breazeal2004humanoid, goodrich2007human}.
Moreover, since people perceive and react to robots differently according to their relationship with robots 
and the appearances of them \cite{sauppe2015social}, it is important to consider these factors
if we want people treat robots as their "work partner" or "friend." 
The LfD paradigm cannot be used to teach robots human-robot collaborative tasks
without considering the above human factors, despite many attractive points of LfD.

There has been a considerable number of previous efforts
to use a LfD paradigm for human-robot collaborative tasks while considering those human-centric issues
using different techniques \cite{nguyen2012capir, dragan2013legibility, cakmak2012designing, ewerton2015learning}.
Even though each work handles the human factors in different ways, very roughly,
we can consider the problems tackled by all of these approaches
as a kind of uncertainty minimization problem
since most of the problems stem from unpredictable behaviors of humans/robots.
Therefore, reducing the uncertainty in communications can be the key concept of these research areas
in the success of the learning process.

For example, many HRI researchers focus more on designing a way to help a human user easily understand their robot partner,
rather than to develop techniques to just transfer knowledge from a human to a robot 
for replicating certain motions \cite{breazeal2004humanoid},
since it helps increase human's predictability.
We as humans tend to feel uncomfortable when we are in unpredictable situations
or when we are not able to understand someone's intention or meaning,
but we would be comfortable as we become more familiar with the situations or each other by understanding them.
As our partner, robots are also required to understand a human partner's mental states
for working together, hence many researchers attempt to automatically detect social cues of 
people using various techniques \cite{calinon2006teaching, misra2016tell}.
We can see details about these research methods in Section 2-A.

Another important perspective to employ the LfD paradigm for human-robot collaborative tasks
is to make the learning process a bidirectional activity rather than 
passive learning where robots learn new tasks from passively observing human demonstrations
\cite{cakmak2012designing}.
This line of research is called \emph{Interactive/Active Learning}
and considers a robot as an active partner that provides feedback to the human teacher
during the collaboration phase, and then the feedback can be used for reducing uncertainty
in terms of accomplishing to learn new tasks so that that learning process is able to become more efficient
\cite{knox2013training}. I will give detailed reviews about the active learning methods in Section 2-B.
%since it helps to increase our predictability.

Learning of complex tasks is another challenge that researchers have recently treated in the field
\cite{figueroa2016learning}. This is one of the ultimate goals of all LfD based approaches
since we want to robots to automatically learn some high-level skills like "pick-up"
without teaching them all the arm trajectories to accomplish the task. 
Although learning of high-level actions is the ultimate goal of the field,
there have been few studies that have investigated it for human-robot collaborative tasks
\cite{mohseni2015interactive}.
I will present these recent studies in Section 2-C.

%Thus, reducing the uncertainty is the key concept of these lines of researches,
%so many HRI researchers more focus on to design a way for making a human user easily understand their robot partner
%rather than to develop techniques to just transfer knowledge from a human to a robot 
%for replicating certain motions \cite{breazeal2004humanoid}.
%As our partner, robots are also required to understand human partner's mental states
%for working together with humans, hence many researchers attempt to detect social cues of 
%people automatically using various techniques \cite{calinon2006teaching, misra2016tell}.

% understanding each ohter mutual process

%In the remaining part of the paper,
%I will review these approaches by splitting into
%three categories according to.

%then uncertainty after they understand each other.
% \cite{lasota2013developing}
%Thus, categorize into three..
%They also tend to view a robot as a partner,
% if we can predictably
%coming from uncertatinty
% can be minimized
%Therefore, many HRI researchers tend to build a framework
%while more focusing on 

% to build framework for helping understanding each other
% can be considered as undertainty mimimization problem

\subsection{Communication}

Communication is the most important key for creating a collaborative robot
that can work around us (humans).
Communication is a bidirectional activity that exchanges mental states
(i.e., thoughts, feelings, and intentions) in both directions,
but researchers in the field tend to focus on one direction
for developing their robotic systems.
However, both are equally important for human-robot collaboration,
I thus introduce recent papers 
that try to make clear communication in both directions
(\emph{``Humans to Robots''} and \emph{``Robots to Humans''})
in this section.

\hfill

\paragraph{User Intention Recognition}
Much of the previous research on LfD for human-robot collaborative tasks
has been carried out to recognize \emph{``Social Cues''}
such as body posture, facial expressions, direction of gaze, and verbal cues in interactions,
since they can give hints to a robot about the goal of the current task for learning.
The robots then are able to use these hints to reduce the search spaces to learn
the new task for speeding up or to correct their movements \cite{billard2008robot}.
Various techniques are used to recognize human user's intentions like
eye-gaze detection \cite{calinon2006teaching}, speech recognition \cite{misra2016tell},
and motion capture \cite{amor2013learning},
but the most intuitive way to let a robot know about our (humans) thoughts or feelings
is probably to use our own (natural) language.

Stefanie Tellex et al. presented a new system for understating human natural language
to automatically generate robot control plans corresponding to the natural language commands
\cite{tellex2011understanding}.
They introduced a new probabilistic graphical model called Generalized Grounding Graphs
in order to transfer a part of natural language control commands to corresponding groundings
like target objects, places, and paths.
The proposed approach is based on Conditional Random Fields (CRFs)
which is one of popular approaches in natural language processing.
They trained the suggested system on a new dataset that they collected from 45 subjects for 22 different videos
using Amazon’s Mechanical Turk (AMT).
This was the annotated dataset of natural language commands that corresponds to correct robot actions.
%After training their system, the system was able to interpret high-level natural language control commands
Finally, they were able to interpret high-level natural language control commands 
such as ``Put the tire pallet on the truck'' after training their system,
and then it generated control plans for the robot.

Recently, Dipendra Misra el al. also introduced a new approach to interpret a user's natural language instructions
to generate robotic actions for manipulation tasks \cite{misra2016tell}.
Their approach considered ambiguity of natural language based instructions
since the same instructions can be interpreted differently according to the current situation of a robot
like its location and the state of the target objects.
For example, an instruction such as ``fill the cup with water'' can be interpreted by the robot
either taking a cup and filling it with water from the tap,
or approaching a refrigerator to take a water bottle out from the fridge at first,
and then pouring water into the cup with the water bottle
according to the environment of the robot.

They handled this ambiguity of robotic instructions and the large variations
given by human natural language
using a CRF based method with a new energy function
that encodes those properties into environment variables for manipulation tasks.
Here, the energy function is composed of several nodes and factors 
that represent natural language (which is converted into a set of verb clauses), environment and controller instruction.
To train this model, they first created a new dataset, Verb-Environment-Instruction Library (VEIL)-300,
which has six different tasks while considering service robot scenarios like
``Making coffee'' or ``Serving affogato.''
The dataset contains natural language commands, environment information, and ground-truth instruction sequences that
correspond to the commands.
They trained their model on this dataset for mapping the natural language commands 
to robot controller instructions, and then they showed accuracy of their model, 61.8\%,
which outperformed all of their baselines on the validation set.

Motion capture is another system that is widely used 
for making demonstration datasets for teaching robots in the field.
Jonas Koenemann et al. presented a real-time motion capture system
that enables a robot to imitate human whole-body motions
\cite{koenemann2014real}.
In this approach, human motions were captured using 
inertial sensors attached to the human body segments with an Xsens MVN motion capture system. 
In order to reduce the computational cost,
they simplified a human model, so they only considered
the positions of the end-effectors (i.e., the position of the hands and feet)
and the position of center of mass
instead of considering a high number of parameters to represent all joint positions of the body.
Then, they applied inverse kinematics to find joint angles
given the positions of end-effectors,
then generated robot motions while considering finding stable robot configurations
instead of just focusing on imitating the human motions directly.
They demonstrated their approach with a Nao humanoid robot,
and then showed the robot was able to imitate human whole-body motions
with consideration for stabilization in real-time.

However, we, as humans, also can catch
other people's thoughts or emotional states from many other signs
like facial expressions, voices or even small body movements
even when we do not understand other people's speaking
or we are not able to see whole-body motions of other people
so these kinds of signs are also very important and meaningful in communication
and widely used by HRI researchers.

Bilge Mutlu showed how embodied cues like facial expressions,
gaze, head gestures, arm gesture, social touch, social smile, etc.
play important roles in communication
through his investigation \cite{mutlu2011designing}.
He explored research on human communication, and then found that
the research provides strong evidence about the hypothesis that embodied cues
help achieve positive social and task outcomes in various domains
of social interaction such as ``Learning and Development'' or ``Motivation, Compliance, and Persuasion''
in human-human communication.
Thus, he finally suggested that HRI researchers should study the most effective ways to use such embodied cues
for designing social robots while considering the relationship between particular embodied cues and outcomes
in order to get similar positive outcomes in human-robot interaction.

%His recent work with Sauppe can be a good example of this to show how important of the embodied cues
%are for designing human collaborative robots \cite{sauppe2015social}.
His recent work with Sauppe can be a good example which shows the importance of embodied cues
for designing human collaborative robots \cite{sauppe2015social}.
In this paper, they studied how robots are treated by human co-workers
in an industrial setting, while focusing on aspects of the robot's design
and context.
The authors found that workers perceive the robots very differently
according to various aspects like the physical appearance of the robots
or their positions (roles) at their places of work.
For example, workers who were supposed to operate the robot treated the robot as their ``work partner'' or ``friend,''
while maintenance and management staff just considered the robot the same as other industrial equipment.
Another interesting finding here is that human workers felt that
the robots had some intelligence because of the robot's eye movements
since it seemed the robots knew what they were doing.
Actually, they were pre-programmed movements so that the robots just moved their eyes
to follow the trajectory of their arms,
however, even though those movements were simple, the movements helped the human workers to understand
the status of the robots and their next actions. Thus, it made human workers feel safe when
they were working in close proximity to the robots, since they believed the robots were able to convey their intentions through the eyes.

% The above paper actually more related to the next section about to 
% show robot's intention

Gesture recognition is also widely used
for human-robot collaboration since gestures can be one of the effective communication channels
between humans and robots for working together \cite{liu2017gesture}.
Various sensors (i.e., a depth camera and a wired glove) and algorithms
(i.e., Hidden Markov Models (HMMs)-based algorithms for modeling meaningful gestures
and skeletal based algorithms for feature extraction and detection)
are used for this line of research.

Jim Mainprice and Dmitry Berenson presented a new framework to recognize human’s intentions as early as
possible \cite{mainprice2013human}.
In this paper, the authors focused on building a framework for early detection of human motion 
in order to generate safe robot motions when humans and robots are working together in close proximity.
They modeled a human's motion as a Gaussians Mixture Model (GMM) representation
and performed Gaussian Mixture Regression (GMR) to predict the human's future motions.
Finally, the proposed approach generated robot motions
while considering a prediction of human workspace occupancy
which is obtained by the swept volume of predicted human motion trajectories.
They demonstrated their approach in a PR2 robot simulation
after training the framework on the collected human motion demonstrations
of manipulation tasks on a table.
It showed that the proposed approach was able to take into account the human motion predictions
in the robot's motion planner, so the robot could interact with the human co-worker more safely and efficiently
in close proximity.

%Anticipating human activities using object affordances for reactive robotic response
Another a interesting keyword that can be used
for understanding user's intention is \emph{``Affordance.''}
Hema S. Koppula and Ashutosh Saxena presented
a new CRF-based approach, called an anticipatory temporal conditional random field (ATCRF),
to predict future human activities based on object affordances \cite{koppula2016anticipating}.
Given the current observation of a human user's pose and the surrounding environment of her/him,
the goal of the proposed approach is to anticipate what the user will do next.
In order to achieve the goal,
they first segmented an observed activity in time,
then constructed a spatio-temporal graph based on the segmented sub-activities.
The graph consists of four types of nodes
(human pose, object affordance, object location, and sub-activity), 
then they augmented the constructed graph with anticipated nodes
representing potential temporal segments. 
The authors demonstrated the proposed approach on
the CAD-120 human activity dataset \cite{koppula2013learning} and obtained 2.7\% improvement
on the state-of-the-art detection results in terms of a success rate.
They also reported that this approach achieved
75.4\%, 69.2\% and 58.1\% accuracy to anticipate
an activity for times of 1, 3 and 10 seconds before, respectively.

Daqing Yi and Michael A. Goodric presented a new framework
for sharing information between robots and humans for task-oriented collaboration \cite{yi2014supporting}.
In this work, they considered a cordon and search mission,
which is one kind of military tactic for searching out the enemy in an area, as a human-robot collaborative task
that has to be solved by a human-robot team.
Here, the authors assumed that a team supervisor (normally a human)
assigns sub-tasks to his/her robot team members
after decomposing the task,
then the robot team members are supposed to accomplish these given sub-tasks
(i.e., searching a high risk sub-region for their human team members). 
They suggested the concept of a shared mental model
for sharing knowledge about the current situation among all team members (robots and humans),
so their framework was presented to help all human and robot team members understand
each other correctly according to their task.
Understanding all commands from natural language is not easy,
but the problem becomes easier in general
if all team members know about the goal,
so in this paper, the authors suggested to use a task-specific (oriented) grammar
for converting a human supervisor's verbal command into a sequence of way points,
thus robot team members could understand their given tasks more correctly.

\hfill

\paragraph{Readable Robot Intentions}
Another direction of LfD research in human-robot collaboration that focuses on 
communication between robots and humans is designing robot behaviors and functions more carefully
in order to show more readable robot intentions.

Breazeal et al. showed that a robot's non-verbal cues are important
for building teamwork between humans and robots \cite{breazeal2005effects}.
In this paper, they recruited a total of 21 subjects then 
conducted a user study with them about task-oriented interactions
between the subjects and the robot Leonardo.
Each subject first was asked to teach Leonardo
the names of three different colored buttons (red, green and blue)
which were located in front of the robot in its workspace,
and then checked to see that the robot knew the names and locations of the buttons.
After that the subject was asked to guide the robot to turn on all of the buttons.
Experimenters recorded videos while the experiment was performed,
and then gave a questionnaire to the subject after the experiment.
After performing behavioral analysis of the videos,
they found that the robot's non-verbal social cues
(e.g., changes of gaze direction and eye blinks)
helped humans read mental states of the robot
and improved human-robot task performance.
The self-report results from the subjects
also suggested that subjects perceived that the robot
was more understandable when the robot 
showed non-verbal behaviors as well as explicitly
using expressive social cues.

Leila Takayama et al. applied animation principles to create readable robot’s behaviors
\cite{takayama2011expressing}.
In this paper, the authors created a robot animation
which shows different robot behaviors according to their hypotheses
(H1: showing forethought before performing an action would improve a robot's readability,
H2: showing a goal-oriented reaction to a task outcome would positively influence people's subjective perceptions of the robot)
and then measured how people described the robot's intentions (before the action is performed)
and how people perceived the robot in terms of some adjectives such as appealing and intelligent
after conducting a video prototyping study with a total of 273 subjects.
They found that people perceived the robot to be more appealing
and their behaviors were more readable
when the robot showed forethought before taking actions.
They also discovered that showing a reaction made people
feel that the robot was more intelligent.
Even though this research is not LfD based,
it shows potential benefits since the animation principles have been verified
and successfully used to make a character by connecting its actions in animations.
Furthermore, HRI researchers can design robot behaviors and test them using animation
instead of building/programming is physical robot to test new designing of robot motions.

%In recent years, there has been an increasing amount of literature on
%to generate readable robot motions.
Here, it is worth noting that the readable robot behaviors
are not always exactly the same as either the optimal behaviors of the robot
to achieve its goal
or expected robot behaviors that we can predict
when we observe the robot operations.

Anca D. Dragan et al. focused on the difference between two types of robot motions
(predictable robot motion and legible motion) \cite{dragan2013legibility}.
They argued that both robot motions are fundamentally different and often show contradictory properties.
Here, the predictable robot motions mean
those that match with expected behaviors of observers (humans).
On the other hand, the legible robot motions 
mean those that convey their intentions of behaviors clearly.
In this research,
the authors formalized legibility and predictability in the context of goal-directed robot motions,
then modeled both robot motions based on a cost optimization function
which is designed in consideration of the principle of rational action.
Finally, they demonstrated that two types of motions were contradictory through their experiments
with three characters (a simulated point robot, the bi-manual robot mobile manipulator (HERB), and a human).
They found that this difference between two properties derived from inferences in opposing directions,
``action-to-goal'' and ``goal-to-action,''
which refer to an observer's ability to answer the questions:
``what is the function of this action?'' when she/he observes ongoing robot actions 
and ``what action would achieve this goal?'' when the observer knows the robot's goal,
respectively.
Their findings through the experiments supported the theory in Psychology
that humans interpret observed behaviors as goal-directed actions.

The same authors studied the effect of \emph{``familiarization''} on the predictability
of robot motion in their follow-up work \cite{dragan2014familiarization}.
This research originated from the idea of
having users learn from robot demonstrations in order to increase
their ability to predict robot motions (familiarization)
because predictability is one of the keys for building collaborative robots
that can work side by side with humans.
This research direction is opposite making robot motions more predictable,
and it gave us valuable insights about building more natural 
human-robot collaboration frameworks.
They used the same methods that were used in their previous work\cite{dragan2014familiarization}
to generate predictable robot motions, and then conducted a new user study
to see the effect of familiarization on the robot motions.
They recruited a total of 50 participants via AMT
and conducted familiarization tests on two different types of robot motion
(natural motion vs. unnatural motion), where the natural motion was defined
as motion that is predictable without (or prior to) familiarization.
In the experiment, each participant was asked to answer the questions
about their predictability of the robot motion
before and after exposing the examples of robot demonstration videos.
They found that the robot motions became
more predictable after familiarization
even though the familiarization was not enough
for users to identify the robot motions, especially
when the robot operated in high-dimensional space with certain complex movements.
The authors also reported that familiarization could help
humans to be more comfortable to robots
and less natural robot motions hindered our ability to predict the motions.

Leah Perlmutter et al. tried to make robots provide their internal states to human users
for helping them to understand the robot's thought and intentions,
since we as humans are not able to judge what robots can see, hear, or infer
in the same way that we use in human-human communication \cite{perlmutter2016rss}.
In this paper, the authors proposed visualization-based
transparency mechanisms rather than developing
a human-like verbal (or non-verbal) communication system for robots.
The proposed visualization module is one kind of the add-on tools
which could be added on a robotic perception system
that consists of three perception components (scene perception, pointing detection, and speech processing)
to interpret the robot user's commands.
They conducted a user study with 20 participants with the proposed robotic system,
and then investigated the effect of their visualization-based transparency mechanisms.
Their findings indicate that visualizations can help users communicate with the robot
and understand robot's abilities
even though some participants reported that they still prefer to have
human-like transparency mechanisms with robots.

\subsection{Interactive/Active Learning}

In recent years,
the research of interactive/active learning has received considerable critical attention in the field.

Maya Cakmak and Andrea L. Thomaz introduced this new robot learning method, which is called \emph{Active Learning},
to allow a robot to ask questions to its teacher (a human user)
when the robot is unsure what to do next during learning \cite{cakmak2012designing}.
In this article, they identified three types of queries
(label, demonstration and feature queries) for an Active Learning based method in LfD
and conducted two sets of experiments with human subjects.
The first set of experiments was designed to investigate how humans ask questions in
human-human collaboration scenarios with some levels of abstraction of the tasks
in consideration of employing the same scenarios for human-robot collaboration.
The second set of experiments was designed to evaluate the use of the three types of queries 
in human-human collaboration scenarios.
The authors found that participants perceived the robot as the smartest
when it asked questions using feature queries,
which is to directly ask about specific features
like positions and rotations to manipulate target objects for learning a new task 
(e.g., ``Should I keep this orientation at the start?'').
They also reported that this type of query 
was the most commonly used in human learning (82\%)
even though this is the most challenging type of query
for robots to produce automatically since 
it requires some level of situation understanding for asking good questions.
These findings provide guidelines to design good questions
for building robots as an active learner in human-robot collaboration scenarios.
%asking behaviors on
%which is the type of queries that 
%were the most common in human learning (82\%

Stefanie Tellex et al. presented an approach for a robot to take advantage of receiving help
from its human partner when the robot and the human partner work together for accomplishing a certain task \cite{tellex2014asking}.
They used a natural language generation system, which is called inverse semantics,
for making a robot that can request help to the human partner in the form of natural language
when the robot fails to do some task,
so that the robot could recover from the failure based on their help.
Since it is impossible to make a perfect robot that never fails,
they focused on developing this recovery method
based on a natural language generation system
for mapping from a desired human helping behavior that the robot would like the human
to execute to words in natural language commands.
This system was then used for generating requests when the robot needs assistance.
When the robot detects failures using a motion capture system (VICON),
their system first represents the failure in a simple symbolic language
which indicates the desired human action,
and then translates this symbolic representation to a natural language sentence using a context free grammar (CFG)
to ask a human for assistance. 
In this research, the authors demonstrated their approach on a human-robot collaborative task of assembling a table together,
and then conducted a user study to evaluate the effectiveness of the proposed approach.
The experimental results showed that the proposed approach
helped the participants infer the requested action from the robot better
than their baselines approaches such as always using a general request (e.g., ``Help me'')
and generating requests using template based methods (e.g., ``Hand me part 2'').
%to improve the speed and accuracy of a human’s intervention. 
%increased the effectiveness of human intervention

W. Bradley Knox et al. presented a case study of teaching a physically embodied robot by human feedback
based on their framework, which is called TAMER (Training an Agent Manually via Evaluative Reinforcement),
that they previously proposed for robot learning from human reward \cite{knox2013training}.
%that they proposed previously for robot learning from human reward \cite{knox2013training}.
In this paper, the authors focused on teaching 
interactive navigation behaviors to their Mobile-Dexterous-Social (MDS) robot Nexi
using human feedback as the only training resource.
There were two buttons for providing positive or negative reward to the robot learner
according to its state-action pair, and robot then was able to be trained given the human reward.
The authors taught a total of five navigation behaviors
such as ``Go to,'' ``Keep conversational distance,'' and ``Look away'' to the robot,
then they tested the learned robot behaviors.
However, they found that Nexi did not move properly after training
due to issues of \emph{transparency}. 
These transparency issues arose due to mismatches between the current state-action pair of the robot learner
and what the human-trainer was observing.
The authors pointed out that there were two main reasons for making this confusion:
1) There can be a delay in the robot taking an action,
so that the mismatch between human observations and internal states of the robot can happen at this point,
2) The perception system of the robot is not perfect,
thus the robot is not able to see some objects around it even if the human trainer can see them.
The authors suggested that researchers should address these transparency challenges
when they employ a human feedback based robot learning method for teaching a physically embodied robot.

%The feedback is provided during learning phase

Karol Hausman et al. presented an approach based on the \emph{interactive perception} paradigm
which uses robot's actuators for actively getting more information about the environment (world)
when the robot is unsure for making a decision at the moment \cite{hausman2015active}. 
They proposed a particle filter-based approach to combine visual robotic perception with
the outcomes of the robot's manipulation actions in a probabilistic way,
and the robot then found the best action to reduce uncertainty over articulated motion models 
given all sensory inputs at the moment.
Here, the articulated motion models indicate the possible movements of objects
such as certain directions (or rotations) of the objects that can be used for manipulating them.
For example, a door of drawers or cabinets has parts that can be moved (also cannot be moved) for opening/closing it
and it can provide useful information to a robot for manipulating the door
since the information can be used for reducing the manipulation space.
In this work, they considered four types of articulated motion models:
rigid, prismatic, rotational and free-body,
and then parametrized them with different numbers of variables according to the types.
They demonstrated the proposed approach using a PR2 mobile manipulator,
and then their experimental results supported that
the robot was able to effectively reduce uncertainty over models
in four manipulation scenarios
(opening and closing of a rotational cabinet door, moving a whiteboard eraser in a straight line,
opening a locked drawer, and grasping a stapler on a table),
and the robot then selected the best action based on a KL-divergence based information gain approach.  

Stefanos Nikolaidis and Julie Shah introduced an interactive training method,
which is called \emph{Cross-training}, for improving human-robot teamwork \cite{nikolaidis2013human}.
A human and a robot are supposed to switch their roles during the training phase
for learning a new collaborative task by cross-training.
This training approach can be considered as a mutual adaptation process.
They reported that a human-robot team performance was significantly improved by cross-training
for accomplish a collaborative task, a simple place-and-drill task,
in their experimental results with human subjects.
The authors also showed that participants who iteratively switched 
their positions with their robot partner, Abbie, perceived 
the robot much more positively than their comparison group
who trained with the robot using standard reinforcement learning methods
in the post experimental survey.
Their findings suggest that
we are able to get better team performance with a robot partner
for accomplishing certain tasks together
when we switch our role with the robot during training phase
in a way similar to human-human team training practices.

\subsection{Learning of Complex Tasks}
Learning complex tasks is
one of the most challenging aspects of employing the LfD paradigm in the field.
%Teaching complex tasks to our robot partner is
%one of the most challenging aspects of employing the LfD paradigm in the field.
Therefore, to date, there are few studies that have investigated
LfD based learning for teaching complex human-robot collaborative tasks to a robot \cite{chernova2014robot}.
Most of them used a decomposing method to make a single complex task
into multiple relatively easy sub-tasks for training a robot.
%One of the possible approaches that can be used for teaching complex tasks to a robot
%is decomposing the complex tasks into a number of single task before performing training.

Scott Niekum et al. presented a Hidden Markov model (HMM) based method,
which is called Beta Process Auto Regressive HMM (BP-AR-HMM), to segment unstructured demonstrations
into multiple sub-skills that enable a robot to learn complex demonstrations in a single integrated framework\cite{niekum2012learning}.
%for enabling a robot to learn multiple sub-tasks in a single integrated framework\cite{niekum2012learning}.
Here, the authors pointed out four key requirements for robot learning of complex tasks:
%to segment unstructured demonstrations into component skills that a robot can learn:
1) the robot must have an ability to recognize repeated instances of skills and generalize them;
2) the robot should be able to do segmentation without prior knowledge;
3) a broad/general class of skills should also be identified by the robot; 
and 4) the robot should be able to represent the skills properly for learning new policies. 
In this paper, they addressed all of the above requirements using BP-AR-HMM 
and Dynamic Movement Primitives (DMPs) which is a framework for representing dynamical systems.
They then demonstrated that the proposed approach helped
the robot learn a multi-step task from unstructured demonstrations.
%generalizing

Nadia Figueroa et al. also employed BP-HMM based approach
for teaching a complex sequential task, pizza dough rolling, to a robot from human demonstrations \cite{figueroa2016learning}.
In this paper, they first extracted a set of unique action primitives
(reach, roll and reach back) and their transition probabilities using an extended version of BP-HMM,
and then trained the model on human demonstrations to learn low-level robot control parameters for
generating proper robot control commands corresponding to each action primitive.
The authors evaluated the proposed framework on the pizza dough rolling task with a real robot
and showed that the robot made the pizza dough with consistent shapes and a desired size
while their baseline approach showed unstable performance in three different types of dough
(very soft, a bit stiffer, and a hard dough) since it used a fixed hand-tuned parameters.

Even though the target scenarios of robot learning in the above mentioned two approaches
are not human-robot collaborative tasks,
both approaches show the challenges for teaching a robot more complex tasks using LfD based approach
and the automatic segmentation methods for handling these issues.
However, in order to apply this line of research for human-robot collaborative tasks,
a robotic learning system should be able to extract different action primitives which are more related to
``interaction'' rather than action itself.

Marco Ewerton et al. presented a Mixture of Interaction Primitives
for learning multiple interaction patterns between two agents (i.e., a human and a robot)
from unlabeled demonstrations \cite{ewerton2015learning}.
Here, Interaction Primitive (IP) is a framework based on DMPs
that was proposed by Heni Ben Amor (who is one of co-authors of this paper)
for robot interactive skill learning and this work is follow-up research
that overcame limits of the previous approach (IP) for learning more complex tasks
and handling various interaction patterns.
The main contribution of this work is modeling multiple interaction patterns
using Gaussian Mixture Model (GMMs) of Interaction Primitives,
so that it enabled modeling nonlinear correlations between
the movements of two different agents (a human and a robot).
In this work, the movements (trajectories) were represented
in the form of the weight vectors, one for each demonstration about a human-robot collaborative task,
and they stacked the several vectors for making a probability distribution.
After that they trained their model to learn interaction patterns
based on the weight vectors that parameterized the trajectories in the demonstrations.
They collected a total of 28 pairs of human-robot demonstrations
for training the proposed framework, and then trained the robot for selecting the appropriate robot reaction
given the observation of the human partner during collaboration.
Their experimental results supported that the robot was able to learn and recognize
multiple human-robot collaborative tasks based on the proposed approach.

\section{Conclusion}
\label{sec:3}

In this paper, I presented a survey of robot learning from demonstration (LfD)
approaches for Human-Robot Collaboration.
LfD is a very attractive research direction
for building a collaborative robot
since it enables robots to automatically learn a new task
from non-robotics experts (end-users).
However, it is also challenging because
there are common theoretical issues like the \emph{Correspondence Problem},
and the situation becomes more challenging when a robot learns complex tasks.
Moreover, researchers should also consider many human-centric issues
such as safety, human partner's feelings, and intention for teaching a robot for collaboration with humans.
Since the human-robot collaborative task is not a manufacturing task that a robot can perform alone (i.e., painting and assembly),
but requires a robot to work side by side as a partner for accomplishing the task,
researchers thus must consider the human-centric issues.

%but researchers should also consider many different aspects of learning for robot collaborative tasks with humans
%rather than other tasks like robot manufacturing tasks that robots can perform alone
%(i.e., painting and assembly) for applying LfD paradigm.
%This is because the human-robot collaborative task is not a task that a robot can perform alone,
%but also it is different from a human-human collaborative task.
%Since a robot would work side by side as a partner for accomplishing the tasks,
%researchers thus need to consider many human-centric issues 
%such as safety, human partner’s feeling and intention for teaching the robot for collaboration.

The most important key word here is communication
because a way to communicate between two agents (a robot and a human) 
can be very different from a way of communication between people,
even though there are a lot of efforts to make this human-robot communication similar to human-human communication.
Communication is a bidirectional activity,
however a lot of robotics researchers tend to view this activity from one direction.
In this survey paper, I thus categorized the LfD based research that focuses on a communication problem
into two lines of works according to their research directions 
and described many interesting findings in the both directions.

Several attempts have been made to use non-verbal cues
such as facial expressions, gaze directions, and body gestures in human-robot communication
since these signals can give out additional useful information for natural and effective communication
between a human and a robot.
In addition, some research has been carried out for designing human readable robot behaviors
(predictable and legible robot motions) for conveying a robot's intentions during collaborations.

Another interesting line of research in the field is the 
works which employ an interactive/active based method for robot learning.
These studies suggest seeing a robot as an active learner that can ask questions
when the robot is unsure what is going on and what to do next.
Furthermore, the robot can actively move itself for gathering more information
for accomplishing/learning a new task in those kinds of situations.

One of the most difficult remaining challenges in the field
is teaching complex collaborative tasks to a robot.
Given a demonstration of complex tasks
(i.e., assembling a table)
that include a number of sub-tasks
(i.e., picking-up a part, holding a part, and turning a screw),
we want the robot to automatically find the sub-tasks,
then generalize and learn them from the demonstrations.
%Learning of complex tasks becomes more challenging
%if we want to teach a robot human-robot collaborative tasks
Since each teacher (a human partner in common) can have different
ways of teaching a robot learner
%ways of teaching and accomplishing tasks with a robot learner
which involve large variations in movements,
teaching a robot human-robot collaborative tasks becomes more challenging.
Hidden Markov Model (HMM) based approaches are widely used
to model dynamics in interaction for learning and they show some possibilities,
but learning complex collaborative tasks still remains open due to the difficulty of the problem,
so researchers, as yet, only consider teaching a relative simple task for robot learning. 

Recently, deep learning based approaches have been widely used in many applications
including object detection, scene segmentation,
and learning robot motor control policy for grasping objects\cite{levine2016learning}.
However, only a few previous studies have investigated applying deep learning based techniques
for teaching robots human-collaborative tasks from demonstrations.
In my view, the main reason is that it is hard to build a large-scale dataset,
which is required for training a robotic system to apply deep learning based methods.
%due to the \emph{Correspondence Problem}.
Different robots have different abilities with different body configurations
and different people want to teach the robots different tasks,
so all of them make the problem harder.

As we see in the paper, most of research in the field still focuses on
making better robotic perception components to understand human's intentions in communication.
In my opinion, we apply deep learning techniques to this line of work without difficulty
then improve an ability of robots to understand human's intentions.
However, making people understand robots is a relatively hard problem
because each robot has its own unique robotic system.
We can teach robots to mimic a human's motion, facial/body expressions for conveying robot's intentions in the same way as humans
using human-human demonstrations in various collaborative scenarios,
but I think that there can be better alternative ways for robots to express their intentions and feelings as robots.
Moreover, even if human-human interactions can give us valuable insights for building robots that can be used in the same interaction scenarios,
simply imitating what humans do may not guarantee the best solution for robot learning of human-robot collaborative tasks.
Since each robot has its unique appearance and functions that are usually quite different from humans,
researchers need to consider how to transfer the learned knowledge from human demonstrations
to each unique intelligent agent.
Consequently, the learned knowledge should be adapted to the robot's unique form.

Many possibilities would be open if we consider robots as our active partners
like interactive/active learning based research,
and then we can take advantage of capabilities of robots themselves for teaching them.
However, one drawback of this line of work is it normally considers that human teachers
exist in the same place for teaching robots (online learning),
but that may not be easy since teaching robots a new task can be a boring and time consuming job.

In this regard, it is time to think about a new learning framework
while considering all of the above mentioned challenges and possibilities.
%For example, we can teach robots using a human-human demonstration data,
%then adjust their behaviors using the active learning based method
%if we can build a new framework that can use both data for training.
As interdisciplinary research, designing robots to collaborate with humans
requires a lot of backgrounds, and researchers thus need to collaborate and work
more closely with other researchers in different fields to provide the new framework
for human-robot collaboration.

%like researches in 
%As an interactive and active agent,
%robots also can teach us while they are learning.

%In this regard, I think it's time to consider designing robot learning method
%for helping people understand robot's intention and feelings

% key -> communication
% bidirectional but most of researches
% only focus on the one-way communications
% Some researches try to take advantage of
% bidrectional things (active learning)
% they used on line learning only

% most of LfD approaches on line learning
% I agree but also offline learning

%These approaches can be categorized into two directions
%according to data source.
%If we collect the data from robots (whether to use an actual robot or use robotic simulator),
%we can minimize the correspondence problem.
%But, it has a limitation in terms of generalization
%since different robots have different body configurations.
%Learned knowledge therefore should be interpreted again
%when we want to apply this knowledge to other robots
%even if they consider exactly the same task.
%Alternatively, the training data can be gathering from human demonstrations,
%however, it also has its own limitations
%since we would face the correspondence problem for teaching robots a new task
%based on human demonstration data.

%In this article,
%I reviewed both types of approaches that are especially designed for
%human-robot collaborative tasks.
%There are pros and cons in the both types of approaches.

\bibliographystyle{IEEEtran}
\bibliography{references}

\begin{thebibliography}{10}
\providecommand{\url}[1]{#1}
\csname url@samestyle\endcsname
\providecommand{\newblock}{\relax}
\providecommand{\bibinfo}[2]{#2}
\providecommand{\BIBentrySTDinterwordspacing}{\spaceskip=0pt\relax}
\providecommand{\BIBentryALTinterwordstretchfactor}{4}
\providecommand{\BIBentryALTinterwordspacing}{\spaceskip=\fontdimen2\font plus
\BIBentryALTinterwordstretchfactor\fontdimen3\font minus
  \fontdimen4\font\relax}
\providecommand{\BIBforeignlanguage}[2]{{%
\expandafter\ifx\csname l@#1\endcsname\relax
\typeout{** WARNING: IEEEtran.bst: No hyphenation pattern has been}%
\typeout{** loaded for the language `#1'. Using the pattern for}%
\typeout{** the default language instead.}%
\else
\language=\csname l@#1\endcsname
\fi
#2}}
\providecommand{\BIBdecl}{\relax}
\BIBdecl

\bibitem{billard2008robot}
A.~Billard, S.~Calinon, R.~Dillmann, and S.~Schaal, ``Robot programming by
  demonstration,'' in \emph{Springer handbook of robotics}.\hskip 1em plus
  0.5em minus 0.4em\relax Springer, 2008, pp. 1371--1394.

\bibitem{argall2009survey}
B.~D. Argall, S.~Chernova, M.~Veloso, and B.~Browning, ``A survey of robot
  learning from demonstration,'' \emph{Robotics and autonomous systems},
  vol.~57, no.~5, pp. 469--483, 2009.

\bibitem{chernova2014robot}
S.~Chernova and A.~L. Thomaz, ``Robot learning from human teachers,''
  \emph{Synthesis Lectures on Artificial Intelligence and Machine Learning},
  vol.~8, no.~3, pp. 1--121, 2014.

\bibitem{nehaniv2002correspondence}
C.~L. Nehaniv, K.~Dautenhahn \emph{et~al.}, ``The correspondence problem,''
  \emph{Imitation in animals and artifacts}, vol.~41, 2002.

\bibitem{yokokohji2005motion}
Y.~Yokokohji, Y.~Kitaoka, and T.~Yoshikawa, ``Motion capture from
  demonstrator's viewpoint and its application to robot teaching,''
  \emph{Journal of Field Robotics}, vol.~22, no.~2, pp. 87--97, 2005.

\bibitem{lasota2013developing}
P.~Lasota, S.~Nikolaidis, and J.~A. Shah, ``Developing an adaptive robotic
  assistant for close proximity human-robot collaboration in space,'' in
  \emph{AIAA Infotech@ Aerospace (I@ A) Conference}, 2013, p. 4806.

\bibitem{breazeal2004humanoid}
C.~Breazeal, A.~Brooks, J.~Gray, G.~Hoffman, C.~Kidd, H.~Lee, J.~Lieberman,
  A.~Lockerd, and D.~Mulanda, ``Humanoid robots as cooperative partners for
  people,'' \emph{Int. Journal of Humanoid Robots}, vol.~1, no.~2, pp. 1--34,
  2004.

\bibitem{goodrich2007human}
M.~A. Goodrich and A.~C. Schultz, ``Human-robot interaction: a survey,''
  \emph{Foundations and trends in human-computer interaction}, vol.~1, no.~3,
  pp. 203--275, 2007.

\bibitem{sauppe2015social}
A.~Saupp{\'e} and B.~Mutlu, ``The social impact of a robot co-worker in
  industrial settings,'' in \emph{Proceedings of the 33rd Annual ACM Conference
  on Human Factors in Computing Systems}.\hskip 1em plus 0.5em minus
  0.4em\relax ACM, 2015, pp. 3613--3622.

\bibitem{nguyen2012capir}
T.-H.~D. Nguyen, D.~Hsu, W.-S. Lee, T.-Y. Leong, L.~P. Kaelbling,
  T.~Lozano-Perez, and A.~H. Grant, ``Capir: Collaborative action planning with
  intention recognition,'' \emph{arXiv preprint arXiv:1206.5928}, 2012.

\bibitem{dragan2013legibility}
A.~D. Dragan, K.~C. Lee, and S.~S. Srinivasa, ``Legibility and predictability
  of robot motion,'' in \emph{ACM/IEEE International Conference on Human-Robot
  Interaction (HRI)}, 2013, pp. 301--308.

\bibitem{cakmak2012designing}
M.~Cakmak and A.~L. Thomaz, ``Designing robot learners that ask good
  questions,'' in \emph{Proceedings of the seventh annual ACM/IEEE
  international conference on Human-Robot Interaction}.\hskip 1em plus 0.5em
  minus 0.4em\relax ACM, 2012, pp. 17--24.

\bibitem{ewerton2015learning}
M.~Ewerton, G.~Neumann, R.~Lioutikov, H.~B. Amor, J.~Peters, and G.~Maeda,
  ``Learning multiple collaborative tasks with a mixture of interaction
  primitives,'' in \emph{IEEE International Conference on Robotics and
  Automation (ICRA)}, 2015, pp. 1535--1542.

\bibitem{calinon2006teaching}
S.~Calinon and A.~Billard, ``Teaching a humanoid robot to recognize and
  reproduce social cues,'' in \emph{The 15th IEEE International Symposium on
  Robot and Human Interactive Communication (RO-MAN)}, 2006, pp. 346--351.

\bibitem{misra2016tell}
D.~K. Misra, J.~Sung, K.~Lee, and A.~Saxena, ``Tell me dave: Context-sensitive
  grounding of natural language to manipulation instructions,'' \emph{The
  International Journal of Robotics Research}, vol.~35, no. 1-3, pp. 281--300,
  2016.

\bibitem{knox2013training}
W.~B. Knox, P.~Stone, and C.~Breazeal, ``Training a robot via human feedback: A
  case study,'' in \emph{International Conference on Social Robotics}.\hskip
  1em plus 0.5em minus 0.4em\relax Springer, 2013, pp. 460--470.

\bibitem{figueroa2016learning}
N.~Figueroa, A.~L. Pais~Ureche, and A.~Billard, ``Learning complex sequential
  tasks from demonstration: A pizza dough rolling case study,'' in \emph{The
  Eleventh ACM/IEEE International Conference on Human Robot Interaction}, 2016,
  pp. 611--612.

\bibitem{mohseni2015interactive}
A.~Mohseni-Kabir, C.~Rich, S.~Chernova, C.~L. Sidner, and D.~Miller,
  ``Interactive hierarchical task learning from a single demonstration,'' in
  \emph{Proceedings of the Tenth Annual ACM/IEEE International Conference on
  Human-Robot Interaction}, 2015, pp. 205--212.

\bibitem{amor2013learning}
H.~B. Amor, D.~Vogt, M.~Ewerton, E.~Berger, B.~Jung, and J.~Peters, ``Learning
  responsive robot behavior by imitation,'' in \emph{IEEE/RSJ International
  Conference on Intelligent Robots and Systems (IROS)}, 2013, pp. 3257--3264.

\bibitem{tellex2011understanding}
S.~A. Tellex, T.~F. Kollar, S.~R. Dickerson, M.~R. Walter, A.~Banerjee,
  S.~Teller, and N.~Roy, ``Understanding natural language commands for robotic
  navigation and mobile manipulation,'' 2011.

\bibitem{koenemann2014real}
J.~Koenemann, F.~Burget, and M.~Bennewitz, ``Real-time imitation of human
  whole-body motions by humanoids,'' in \emph{IEEE International Conference on
  Robotics and Automation (ICRA)}, 2014, pp. 2806--2812.

\bibitem{mutlu2011designing}
B.~Mutlu, ``Designing embodied cues for dialog with robots,'' \emph{AI
  Magazine}, vol.~32, no.~4, pp. 17--30, 2011.

\bibitem{liu2017gesture}
H.~Liu and L.~Wang, ``Gesture recognition for human-robot collaboration: A
  review,'' \emph{International Journal of Industrial Ergonomics}, 2017.

\bibitem{mainprice2013human}
J.~Mainprice and D.~Berenson, ``Human-robot collaborative manipulation planning
  using early prediction of human motion,'' in \emph{IEEE/RSJ International
  Conference on Intelligent Robots and Systems (IROS)}, 2013, pp. 299--306.

\bibitem{koppula2016anticipating}
H.~S. Koppula and A.~Saxena, ``Anticipating human activities using object
  affordances for reactive robotic response,'' \emph{IEEE transactions on
  pattern analysis and machine intelligence}, vol.~38, no.~1, pp. 14--29, 2016.

\bibitem{koppula2013learning}
H.~S. Koppula, R.~Gupta, and A.~Saxena, ``Learning human activities and object
  affordances from rgb-d videos,'' \emph{The International Journal of Robotics
  Research}, vol.~32, no.~8, pp. 951--970, 2013.

\bibitem{yi2014supporting}
D.~Yi and M.~A. Goodrich, ``Supporting task-oriented collaboration in
  human-robot teams using semantic-based path planning,'' in \emph{Proc. SPIE},
  vol. 9084, 2014.

\bibitem{breazeal2005effects}
C.~Breazeal, C.~D. Kidd, A.~L. Thomaz, G.~Hoffman, and M.~Berlin, ``Effects of
  nonverbal communication on efficiency and robustness in human-robot
  teamwork,'' in \emph{IEEE/RSJ International Conference on Intelligent Robots
  and Systems (IROS)}, 2005, pp. 708--713.

\bibitem{takayama2011expressing}
L.~Takayama, D.~Dooley, and W.~Ju, ``Expressing thought: improving robot
  readability with animation principles,'' in \emph{Proceedings of the 6th
  international conference on Human-robot interaction}.\hskip 1em plus 0.5em
  minus 0.4em\relax ACM, 2011, pp. 69--76.

\bibitem{dragan2014familiarization}
A.~Dragan and S.~Srinivasa, ``Familiarization to robot motion,'' in
  \emph{Proceedings of the 2014 ACM/IEEE international conference on
  Human-robot interaction}.\hskip 1em plus 0.5em minus 0.4em\relax ACM, 2014,
  pp. 366--373.

\bibitem{perlmutter2016rss}
L.~Perlmutter, E.~Kernfeld, and M.~Cakmak, ``Situated language understanding
  with human-like and visualization-based transparency,'' 2016.

\bibitem{tellex2014asking}
S.~Tellex, R.~A. Knepper, A.~Li, D.~Rus, and N.~Roy, ``Asking for help using
  inverse semantics.'' in \emph{Robotics: Science and systems}, 2014.

\bibitem{hausman2015active}
K.~Hausman, S.~Niekum, S.~Osentoski, and G.~S. Sukhatme, ``Active articulation
  model estimation through interactive perception,'' in \emph{IEEE
  International Conference on Robotics and Automation (ICRA)}, 2015, pp.
  3305--3312.

\bibitem{nikolaidis2013human}
S.~Nikolaidis and J.~Shah, ``Human-robot cross-training: computational
  formulation, modeling and evaluation of a human team training strategy,'' in
  \emph{Proceedings of the 8th ACM/IEEE international conference on Human-robot
  interaction}, 2013, pp. 33--40.

\bibitem{niekum2012learning}
S.~Niekum, S.~Osentoski, G.~Konidaris, and A.~G. Barto, ``Learning and
  generalization of complex tasks from unstructured demonstrations,'' in
  \emph{IEEE/RSJ International Conference on Intelligent Robots and Systems
  (IROS)}, 2012, pp. 5239--5246.

\bibitem{levine2016learning}
S.~Levine, P.~Pastor, A.~Krizhevsky, and D.~Quillen, ``Learning hand-eye
  coordination for robotic grasping with deep learning and large-scale data
  collection,'' \emph{arXiv preprint arXiv:1603.02199}, 2016.

\end{thebibliography}

\end{document}